\documentclass[11pt]{article}
\pdfoutput=1

\usepackage{fullpage,times}

\usepackage[utf8]{inputenc}
\usepackage[T1]{fontenc}
\usepackage{hyperref}
\hypersetup{
        colorlinks   = true,
        urlcolor     = blue,
        linkcolor    = blue,
        citecolor   = blue
}

\usepackage{url}
\usepackage{booktabs}
\usepackage{amsmath,amsfonts,amsthm,amssymb}
\usepackage{nicefrac}
\usepackage{microtype}

\usepackage[round]{natbib}

\usepackage{tikz}
\usepackage{pgfplots}
\pgfplotsset{compat=1.8} 
\usepackage{tkz-graph}

\usepackage{enumitem}
\usepackage[nolist,nohyperlinks]{acronym}
\usepackage{color}
\usepackage{bold-extra}
\usepackage{mathtools}
\usepackage{etoolbox}
\usepackage{caption}
\usepackage{subcaption}
\usepackage[utf8]{inputenc}
\usepackage{amsthm} 
\usepackage{amsmath}
\usepackage{amsfonts}
\usepackage{mathtools} 
\usepackage{amsmath,amsfonts,amsthm,amssymb}
\usepackage{algorithm}
\usepackage{algpseudocode}
\usepackage{graphicx}
\graphicspath{{plots/}}
\usepackage{authblk}

\newcommand{\sA}{\mathcal{A}}

\newcommand{\scK}{\mathcal{K}}

\DeclareMathOperator*{\argmax}{arg\,max}

\newcommand{\field}[1]{\mathbb{#1}}

\newcommand{\Nat}{\field{N}}

\newcommand{\btau}{\boldsymbol{\tau}}

\newcommand{\theset}[2]{ \left\{ {#1} \,:\, {#2} \right\} }

\newcommand{\scO}{\mathcal{O}}

\newcommand{\dt}{\displaystyle}

\newcommand{\wh}{\widehat}
\newcommand{\ve}{\varepsilon}

\newtheorem{theorem}{Theorem}

\newtheorem{fact}{Fact}

\DeclareMathOperator*{\E}{\mathbb{E}}

\newtheoremstyle{named}{}{}{\itshape}{}{\bfseries}{}{.5em}{\thmnote{#3}#1}
\theoremstyle{named}

\newcommand{\muhat}{\wh{\mu}}
\newcommand{\ghat}{\wh{g}}
\newcommand{\mhat}{\wh{m}}

\newcommand{\mubar}{\mu}
\newcommand{\dbar}{d}

\DeclarePairedDelimiter\ceil{\lceil}{\rceil}

\renewcommand{\Pr}{\field{P}}

\newcommand{\Ind}[1]{ \field{I}\left\{{#1}\right\} }

\newcommand{\greedy}{\ensuremath{\pi_{\mathrm{greedy}}}}
\newcommand{\ghost}{\ensuremath{\pi_{\mathrm{ghost}}}}
\newcommand{\low}{\ensuremath{\pi_{\mathrm{low}}}}
\newcommand{\piucb}{\ensuremath{\pi_{\mathrm{ucb}}}}
\newcommand{\rstar}{r^{\star}}

\begin{acronym}
\acro{IC}{Improvement Condition}
\acro{RLS}{Regularized Least Squares}
\acro{TL}{Transfer Learning}
\acro{HTL}{Hypothesis Transfer Learning}
\acro{ERM}{Empirical Risk Minimization}
\acro{TEAM}{Target Empirical Accuracy Maximization}
\acro{RKHS}{Reproducing kernel Hilbert space}
\acro{DA}{Domain Adaptation}
\acro{LOO}{Leave-One-Out}
\acro{HP}{High Probability}
\acro{RSS}{Regularized Subset Selection}
\acro{FR}{Forward Regression}
\acro{PSD}{Positive Semi-Definite}
\acro{SGD}{Stochastic Gradient Descent}
\acro{OGD}{Online Gradient Descent}
\acro{EWA}{Exponentially Weighted Average}
\acro{EMD}{Effective Metric Dimension}
\acro{PDE}{Partial Differential Equation}
\acro{SDE}{Stochastic Differential Equation}
\acro{FD}{Frequent Directions}
\acro{OFU}{Optimism in the Face of Uncertainty}
\acro{TS}{Thompson Sampling}
\end{acronym}

\title{Stochastic Bandits with Delay-Dependent Payoffs}

\author{Leonardo Cella \thanks{leonardocella@gmail.com} }
\author{Nicol\`{o} Cesa-Bianchi \thanks{nicolo.cesa-bianchi@unimi.it}}

\affil{
  Dipartimento di Informatica\\
  Universit\`{a} degli Studi di Milano\\
  20133 Milano, Italy}

\makeindex
\begin{document}
  
\maketitle

\begin{abstract}
Motivated by recommendation problems in music streaming platforms, we propose a nonstationary stochastic bandit model in which the expected reward of an arm depends on the number of rounds that have passed since the arm was last pulled. After proving that finding an optimal policy is NP-hard even when all model parameters are known, we introduce a class of ranking policies provably approximating, to within a constant factor, the expected reward of the optimal policy. We show an algorithm whose regret with respect to the best ranking policy is bounded by $\widetilde{\scO}\big(\!\sqrt{kT}\big)$, where $k$ is the number of arms and $T$ is time. Our algorithm uses only $\scO\big(k\ln\ln T)$ switches, which helps when switching between policies is costly. As constructing the class of learning policies requires ordering the arms according to their expectations, we also bound the number of pulls required to do so. Finally, we run experiments to compare our algorithm against UCB on different problem instances.
\end{abstract}

\section{Introduction}
\label{Sec:intro}
Multiarmed bandits ---see, e.g., \citep{RegretBandits}--- are a popular mathematical framework for modeling sequential decision problems in the presence of partial feedback; typical application domains include clinical trials, online advertising, and product recommendation. Consider for example the task of learning the genre of songs most liked by a given user of a music streaming platform. Each song genre is viewed as an arm of a bandit problem associated with the user. A bandit algorithm learns by sequentially choosing arms (i.e., recommending songs) and observing the resulting payoff (i.e., whether the user liked the song). The payoff is used by the algorithm to refine its recommendation policy. The distinctive feature of bandits is that, after each recommendation, the algorithm gets only a feedback for the selected arm (i.e., the single genre that was recommended).

In the simplest stochastic bandit framework \citep{lai1985asymptotically} rewards are realizations of i.i.d.\ draws from fixed and unknown distributions associated to each arm. In this setting the optimal policy is to consistently recommend the arm with the highest reward expectation. On the other hand, in scenarios like song recommendation, users may grow tired of listening to the same music genre over and over.
This is naturally formalized as a nonstationary bandit setting, where the payoff of an arm grows with the time since the arm was last played. In this case policies consistently recommending the same arm are seldom optimal. E-learning applications, where arms corresponds to questions that students have to answer, are other natural examples of the same phenomenon, as asking again immediately the same question that the student has just answered is not very effective.

In this paper we introduce a simple nonstationary stochastic bandit model, B2DEP, in which the expected reward $\mu_i(\tau)$ of an arm $i$ is a bounded nondecreasing function of the number $\tau$ of rounds that have passed since the arm was last selected by the policy. More specifically, we assume each arm $i$ has an unknown baseline payoff expectation $\mu_i$ (equal to the expected payoff when the arm is pulled for the first time) and an unknown delay parameter $d_i > 0$. If the arm was pulled recently (that is, $1 \le \tau \le d_i$), then the expected payoff may be smaller that its baseline value: $\mu_i(\tau) \le \mu_i$. Vice versa, if $\tau > d_i$, then $\mu_i(\tau)$ is guaranteed to match the baseline value $\mu_i$. In the song recommendation example, the delays $d_i$ model the extent to which listening to a song of genre $i$ affects how much a user is willing to listen to more songs of that same genre.

Since $\tau$ can be viewed as a notion of state for arm $i$, our model can be compared to nonstationary models, such as rested bandits \citep{gittins1979bandit} and restless bandits \citep{whittle1988restless} ---see also \citep{TekinRestedRestless}. In restless bandits the reward distribution of an arm changes irrespective of the policy being used, whereas in rested bandits the distribution changes only when the arm is selected by the policy. Our setting is neither rested nor restless, as our reward distributions change differently according to whether the arm is selected by the policy or not.

In Section~\ref{Sec:Hardness} we make a reduction to the Periodic Maintenance Scheduling Problem \citep{hardScheduling} to prove that the optimization problem of finding an optimal periodic policy in our setting is NP-Hard. In order to circumvent the hardness of computing the optimal periodic policy, in Section~\ref{Sec:RegretDec} we identify a simple class of periodic policies that are efficiently learnable, and whose expected reward is provably to within a constant factor of that of the optimal policy. Our approximating class is pretty natural: it contains all \textsl{ranking policies} that cycle over the $r$ best arms (where $r$ is the parameter to optimize) according to the unknown ordering based on the arms' baseline payoff expectations. 
As it turns out, learning the best ranking policy can be formulated in terms of minimizing the standard notion of regret. This is unlike the problem of learning the best periodic policy, which instead requires minimizing the harder notion of policy regret \citep{arora2012online}.

Consider the task of learning the best ranking policy. In our music streaming example, a ranking policy is a playlist for the user. As changing the playlist streamed to the user may be costly in practice, we also introduce a \textsl{switching cost} for selecting a different ranking policy. 
Controlling the number of switches could also have a good effect in our nonstationary setting, when the expected reward of a ranking policy may depend on which other ranking polices were played earlier. The learning agent should ensure that a ranking policy is played many times consecutively (i.e., infrequent switches), so that estimates are calibrated (i.e., computed in the same context of past plays).

A standard bandit strategy like UCB \citep{UCB1}, which guarantees a regret of $\scO\big(\sqrt{kT\ln T}\big)$ irrespective of the size of the suboptimality gaps between the expected reward of the optimal ranking policy and that of the other policies, performs a number of switches growing with the squared inverse of these gaps. In Section~\ref{Sec:MaxRankStage} we show how to learn the best ranking policy using a simple variant of a learning algorithm based on action elimination proposed in~\citep{cesa2013online}. Similarly to UCB, this algorithm has a distribution-free regret bound of bounded by $\sqrt{kT}$. However, a bound $\scO\big(k\ln\ln T\big)$ on the number of switches is also guaranteed irrespective of the size of the gaps. 

In Section~\ref{Sec:sampling_stage} we turn to the problem of constructing the class of ranking policies, which amounts to learning the ordering of the arms according to their baseline payoff expectations $\mu_1,\ldots,\mu_k$. Assuming $\mu_1 > \cdots > \mu_k$, this can be reduced to the problem of learning the ordering of reward expectations in a standard stochastic bandit with i.i.d.\ rewards. We show that this is possible with a number of pulls bounded by
$
	\sum_i {1}/{\Delta_i^2}
$
(ignoring logarithmic factors), where $\Delta_i$ is the smallest gap between $\mu_{i-1}-\mu_i$ and $\mu_i-\mu_{i-1}$. Note that this bound is not significantly improvable, because $1\big/\Delta_i^2$ samples of arm each $i$ are needed to verify that $\mu_{i-1} < \mu_i < \mu_{i+1}$.

Finally, in Section~\ref{Sec:Experiments} we describe experiments comparing our low-switch algorithm against UCB in both large-gap and in small-gap settings.
\section{Related works}
Our setting is a variant of the model introduced by \citet{kleinberg2018recharging}. In that work, $\mu_i(\tau)$ are concave, nondecreasing functions satisfying $\mu_i (\tau) \le \mu_i(\tau-1) + 1$. Note that this setting and ours are incomparable. Indeed, unlike \citep{kleinberg2018recharging} we assume a specific parametric form for the functions $\mu_i(\cdot)$, which are nondecreasing and bounded by $1$. On the other hand, we do not assume concavity, which plays a key role in their analysis.

\citet{RecoveringBandits} consider a setting in which the expected reward functions $\mu_i(\cdot)$ are sampled from a Gaussian Process with known kernel. The main result is a bound of order $\sqrt{kT}$ on the Bayesian $d$-step lookahead regret, where $d$ is a user-defined parameter. This notion of regret is defined by dividing time in length-$d$ blocks, and then summing the regret in each block against the greedy algorithm optimizing the next $d$ pulls given the agent's current configuration of delays (i.e., how long ago each arm was last pulled). Similarly to \citep{RecoveringBandits}, we also compete against a greedy block strategy. However, in our case the block length is unknown, and the greedy strategy is not defined in terms of the agent's delay configuration.

A special case of our model is investigated in the very recent work by \citet{blockingBandits}. Unlike B2DEP, they assume $\mu_i(\tau) = 0$ for all $\tau \le d_i$ and complete knowledge of the delays $d_i$. In fact, they even assume that every arm $i$ cannot be selected in the next $d_i$ time steps after a pull. Their main result is a regret bound for a variant of UCB competing against the greedy policy. They also show NP-hardness of finding the optimal policy through a reduction similar to ours. It is not clear how their learning approach could be extended to prove results in our more general setting, where $\mu_i(\tau)$ could be positive even when $\tau \le d_i$ and the delays $d_i$ are unknown. 

A different approach to nonstationary bandits in recommender systems considers expected reward functions that depend on the number of times the arm was played so far, \citep{levine2017rotting,cortes2017discrepancy,bouneffouf2016multi,Heidari_DecayingBandits,seznec2019rotting,linUCRL}. These cases correspond to a rested bandit model, where each arm's expected reward can only change when the arm is played.

The fact that we learn ranking strategies is reminiscent of stochastic combinatorial semi-bandits \citep{kveton2015tight}, where the number of arms in the schedule is a parameter of the learning problem. In particular, similarly to \citep{radlinski2008learning, kveton2015cascading, katariya2016dcm} our strategies learn rankings of the actions, but unlike those approaches in our case the optimal number of elements in the ranking must be learned too.

\section{The B2DEP setting}
\label{Sec:Model_def}
In the classical stochastic multiarmed bandit model, at each round $t=1,2,\ldots$ the agent pulls an arm from $\mathcal{K}=\{1,\dots,k\}$ and receives the associated payoff, which is a $[0,1]$-valued random variable independently drawn from the (fixed but unknown) probability distribution associated with the pulled arm. The payoff is the only feedback revealed to the agent at each round. The agent's goal is to maximize the expected cumulative payoff over any number of rounds.

In the B2DEP (Bandits with DElay DEpendend Payoff) variant introduced here, when the agent plays an arm $i \in \mathcal{K}$ the $[0,1]$-valued payoff has expected value
\begin{equation} \label{Eq:Payout}
    \mu_i(\tau) = \left(1 - f(\tau) \mathbb{I}\left\{0 < \tau \leq \dbar_i\right\}\right)\mubar_i
\end{equation}
where $\mubar_i$ is the unknown \textsl{baseline reward expectation} for arm $i$, $f : \Nat \to [0,1]$ is an unknown nonincreasing function, and $\tau$ is the number of rounds that have passed since that arm was last pulled (conventionally, $\tau=0$ means that an arm is pulled for the first time). When $f$ is identically zero, B2DEP reduces to the standard stochastic bandit model with payoff expectations $\mubar_1,\ldots,\mubar_k$. The unknown arm-dependent delay parameters $\dbar_i > 0$ control the number of rounds after which the arm's expected payoff is guaranteed to return to its baseline value $\mubar_i$.

A policy $\pi$ maps a sequence of past observed payoffs to the index of the next arm to pull. Let $g_t(\pi)$ be the payoff collected by policy $\pi$ at round $t$. Given an instance of B2DEP, the optimal policy $\pi^*$ maximizes, over all policies $\pi$, the long term expected average payoff
\[
	\lim_{T\to\infty} \frac{G_T(\pi)}{T} \quad\text{where}\quad G_T(\pi) = \E\left[\sum_{t=1}^T g_t(\pi)\right].
\]
Note that, the payoff expectations at any time step $t$ are fully determined by the current \textsl{delay vector} $\btau(t) = \big(\btau_1(t),\dots,\btau_k(t)\big)$, where each integer $0 \le \tau_i(t) \le \dbar_i$ counts how many rounds have passed since $i\in\scK$ was last pulled (setting $\tau_i(t)=0$ if $i$ was never pulled or if it was last pulled more than $d_i$ steps ago). Hence, any delay-based policy ---e.g., any deterministic function of the current delay vector--- is eventually periodic, meaning that $\pi\big(\btau(t)\big) = \pi\big(\btau(t+P)\big)$ for all $t_0 \le t \le T$, where $P$ is the period and $t_0$ is the length of the transient.


Consider the greedy policy \greedy\ defined as follows: At each round $t$, \greedy\ pulls the arm $i\in\mathcal{K}$ with the highest expected reward according to current delays
\begin{equation}\label{Eq:GreedyPolicy}
	\greedy\big(\btau(t)\big) = \argmax_{i\in\mathcal{K}}\mu_i\big(\tau_i(t)\big)
\end{equation}
where $\tau_i(t)=0$ if $i$ was never pulled before. It is easy to see that \greedy\ is not always optimal. For example consider the following instance of $B2DEP$ with $k=2$: $f(\tau) = \frac{1}{2}$ for all $\tau$, $\mu_1 = 1$, $\mu_2 = \frac{1}{2}-\ve$, $d_1 = d_2 = 1$. Then \greedy\ always pulls arm $1$ and achieves $G_t(\greedy) = 1 + \frac{T-1}{2}$, whereas $G_T(\pi^*) = 1 + \frac{T-1}{2}\big(\frac{3}{2}-\ve \big)$ where $\pi^*$ alternates between arm $1$ and arm $2$. Hence $G_T(\greedy) \le \frac{2}{3}G_T(\pi^*)$.

In the next section we show that the problem of finding the optimal periodic policy for B2DEP is intractable.

\section{Hardness results}
\label{Sec:Hardness}
We show that the optimization problem of finding an optimal policy for B2DEP is NP-hard, even when all the instance parameters are known. Our proof relies on the NP-completeness of the Periodic Maintenance Scheduling Problem (PMSP) shown by \citet{hardScheduling}. Although a very similar result can also be proven using the reduction of \citet{blockingBandits}, introduced for a special case of our B2DEP setting, we give our proof for completeness.

A maintenance schedule on $n$ machines $\{1,\dots,n\}$ is any infinite sequence over $\{0,1,\dots,n\}$, where $0$ indicates that no machine is scheduled for service at that time. An instance of the PMSP decision problem is given by integer service intervals $\ell_1,\dots,\ell_n$ such that $\sum_{i=1}^n \frac{1}{\ell_i} \le 1$. The question is whether there exists a maintenance schedule such that the consecutive service times of each machine $i$ are exactly $\ell_i$ times apart.
The following result holds (proof in the supplementary material).
\begin{theorem}\label{Th:Hardness}
It is NP-hard to decide whether an instance of B2DEP has a periodic policy $\pi$ achieving
\[
	\lim_{T\to\infty} \frac{G_T(\pi)}{T} \ge \sum_{i=1}^k \frac{\mubar_i}{\dbar_i+1}~.
\]
\end{theorem}

\section{Approximating the optimal policy}
\label{Sec:RegretDec}
In order circumvent the computational problem of finding the best periodic policy, we introduce a simple class $\Pi_{\scK}$ of periodic \textsl{ranking policies} whose best element \ghost\ has a cumulative expected payoff not too far from that of $\pi^*$. Without loss of generality, assume that $\mubar_1 > \cdots > \mubar_k$. Let $\Pi_{\scK} \equiv \theset{\pi_m}{m\in\scK}$, where each policy $\pi_m$ cycles over the arm sequence $1,\ldots,m$. The average reward $g(m)$ of policy $\pi_m$ is defined by
\[
	g(m) = \frac{1}{m}\sum_{j=1}^{m} \mu_{j}(m)~.
\]
Since \ghost\ maximizes $g(m)$ over $m\in\scK$, $\ghost \equiv \pi_{\rstar}$ where
\begin{equation}
\label{Eq:n_star}
	\rstar \in \argmax_{m=1,\ldots,k} \frac{1}{m}\sum_{j=1}^{m} \mu_{j}(m)
\end{equation}
We now bound $G_T(\ghost)$ in terms of $G_T(\pi^*)$.
\begin{theorem}
\label{Th:RegretDecomposition}
\[
     G_T(\ghost)
\ge
	 \big(1-f(r_0)\big)G_T(\pi^*) + O(1)
\]
where $r_0$ is the largest arm index $r$ such that
\[
	\mubar_i > \max_{j=1,\ldots,i-1}\mu_j(i-j) \quad i=2,\ldots,r
\]
and $r_0=1$ if $\mubar_2 \le \mu_1(1)$. 
\end{theorem}
The definition of $r_0$ is better understood in the context of the more intuitive delay-based policy \greedy. Note indeed that $r_0+1$ is the first round in which \greedy\ prefers to pull one of the arms that were played in the first $r_0$ rounds rather than the next arm $r_0+1$. 
\begin{proof}
Since $\rstar$ maximizes~(\ref{Eq:n_star}),
\begin{align*}
	G_T(\ghost)
&=
	\frac{T}{\rstar} \sum_{i=1}^{\rstar} \mu_i(\rstar) + \scO(1)
\\ &\ge
	\frac{T}{r_0} \sum_{i=1}^{r_0} \mu_i(r_0) + \scO(1)
\\ &\ge
	\frac{T}{r_0} \sum_{i=1}^{r_0}\big(1-f(r_0)\big) \mubar_i + \scO(1)
\end{align*}
where the $\scO(1)$ term takes into account that $\rstar$ may not divide $T$, and the fact that in the first $\rstar$ rounds the expected reward is $\mubar_1+\cdots+\mubar_{\rstar}$ instead of $\mu_1(\rstar)+\cdots+\mu_{\rstar}(\rstar)$. Now split the $T$ time steps in blocks of length $r_0$. Because $r_0$ is ---by definition--- the largest expected reward any policy can achieve in $r_0$ consecutive steps, the expected reward of $\pi^*$ in any of these blocks is at most $\mu_1+\cdots+\mu_{r_0}$. Therefore 
\[
	G_T(\pi^*) \le \frac{T}{r_0} \sum_{i=1}^{r_0} \mubar_i + \scO(1)
\]
where, as before, the $\scO(1)$ term takes into account that $r_0$ may not divide $T$. This concludes the proof.
\end{proof}
The proof of Theorem~\ref{Th:RegretDecomposition} actually shows that both $\rstar$ and $r_0$ achieve the claimed approximation. However, by definition $G_T(\ghost)$ is bigger than the total reward of the policy that cycles over $1,\ldots,r_0$. Also, learning $\ghost$ is relatively easy, as we show in Section~\ref{Sec:MaxRankStage}.

It is easy to see that $g(m)$ is not monotone due to the presence of the coefficients $\dbar_i$. For example, consider the B2DEP instance defined by $k=3$, $\mubar_1=1$, $\mubar_2=\frac{2}{3}$, $\mubar_3=\frac{1}{2}$, $\dbar_1=\dbar_2=\dbar_3=2$, and $f(\tau) = 2^{-\tau}$. Then $g(2) < g(1) < g(3)$.

\section{Learning the ghost policy}
\label{Sec:MaxRankStage}
In this section we deal with the problem of learning $\rstar$ assuming the correct ordering $1,\ldots,k$ of the arms (such that $\mu_1 > \cdots > \mu_k$) is known. In the next section, we consider the problem of learning this ordering.

Our search space is the set of ranking policies $\Pi_{\scK} \equiv \theset{\pi_m}{m\in\scK}$, where each policy $\pi_m$ cycles over the arm sequence $1,\ldots,m$. Note that, by definition, $\ghost \equiv \pi_{\rstar}$. The average reward $g(m)$ of policy $\pi_m$ is defined by
$
	g(m) = \big(\mu_1(m)+\cdots+\mu_m(m)\big)\big/m
$.
Note that every time the learning algorithm chooses to play a different policy $\pi_m\in\Pi_{\scK}$, an extra cost is incurred due to the need of calibrating the estimates for $g(m)$. In fact, if we played a policy different from $\pi_m$ in the previous round, the reward expectation associated with the play of $\pi_m$ in the current round is potentially different from $g(m)$. This is due to the fact that we cannot guarantee that each arm in the schedule used by $\pi_m$ was pulled exactly $m$ steps earlier. This implies that we need to play each newly selected policy more than once, as the first play cannot be used to reliably estimate $g(m)$.

We now introduce the policy \low\ (Algorithm~\ref{Alg:LowSwitch}), a simple variant of a learning algorithm based on action elimination proposed in ~\citep{cesa2013online}. This policy has a regret bound similar to UCB while guaranteeing a bound $\scO\big(k\ln\ln T\big)$ on the number of switches, irrespective of the size of the gaps. In Section~\ref{Sec:Experiments} we compare \low with UCB.
\begin{algorithm}[!t]
\begin{algorithmic}[1]
\Require{Policy set $\Pi_{\scK}$, confidence $\delta \in (0,1)$, horizon $T$}
\State Let $\sA_1 \equiv \scK$ be the initial set of active policies
\Repeat  \Comment{$s$ indexes the stage number}
	\For {$m \in \sA_s$} \label{line:forloop}
		\State Play $\pi_m$ for $T_s/\big(m|\sA_s|\big) + 1$ times
		\State Compute $\ghat_{s}(m)$ discarding the first play
	\EndFor \label{line:endloop}
	\State Let ${\dt \mhat_s = \argmax_{m\in\sA_s} \ghat_s(\mhat_s) }$ \label{line:empopt}
	\State $\sA_{s+1} = \theset{m\in\sA_s}{\ghat_{s}(m) \ge \ghat_{s}(\mhat_s) - 2C_s}$ \label{line:updateactive}
\Until{overall number of pulls exceeds $T$}
\end{algorithmic}
\caption{(\low)}
\label{Alg:LowSwitch}
\end{algorithm}

In each stage $s$, algorithm \low\ plays each policy $\pi_m$ in the active set $\sA_s$ for $T_s/\big(m|\sA_s|\big) + 1$ times, where $T_s = T^{1-2^{-s}}$. Then, the algorithm computes the sample average reward $\ghat_{s}(m)$ based on these plays, excluding the first one because of calibration (lines~\ref{line:forloop}--\ref{line:endloop}). After that, the empirically best policy is selected (\ref{line:empopt}). Finally, the active set is recomputed (line~\ref{line:updateactive}) excluding all policies whose sample average reward is significantly smaller than that of the empirically best policy. The quantity $C_s$ is derived from a standard Chernoff-Hoeffding bound and is equal to $\sqrt{\frac{k}{2T_s}\ln\frac{2kS}{\delta}}$ where
\begin{align*}
	S = \min\theset{j\in\Nat}{\sum_{s=1}^j \big(|\sA_s| + T_s\big) \ge T}
\end{align*}
implying $S= \scO\big(\ln\ln T\big)$. The terms $|\sA_s|$ account for the extra calibration pull each time we switch to a new policy in $\Pi_{\scK}$. We can prove the following bound on the regret of \low\ with respect to \ghost.
\begin{theorem}
\label{th:lowswitch}
When run on an instance of B2DEP with parameters $\delta$ and $T$, with probability at least $1-\delta$ Algorithm~\ref{Alg:LowSwitch} guarantees
\begin{align}
\nonumber
	G_T&(\ghost) - G_T(\low)
\\ &=
\label{eq:switchbound}
	\scO\left(k^2\ln\ln T + \sqrt{kT\left(\ln\frac{k}{\delta}+ \ln\ln\ln T\right)}\right)
\end{align}
with probability at least $1-\delta$.
\end{theorem}
Note that this bound is distribution-free. That is, it does not depend on the gaps $g(\rstar) - g(m)$ (which in general could be arbitrarily small). The rate $\sqrt{T}$, as opposed to the $\ln T$ rate of distribution-dependent bounds, cannot be improved upon in general \cite{RegretBandits}.

\begin{proof}
The proof is an adaptation of \citep[Theorem 6]{cesa2013online}.
Note that $\sA_S \subseteq \cdots \subseteq \sA_1$ by construction. Also, our choice of $C_s$ and Chernoff-Hoeffding bound implies that
\begin{equation}
\label{eq:uniform}
    \max_{m \in \sA_s} \bigl| \ghat_{s}(m) - g(m) \bigr| \le C_s
\end{equation}
simultaneously for all $s=1,\dots,S$ with probability at least $1-\delta$. To see this, note that in every stage $s$ the estimates $\ghat_s(m)$ are computed using $T_s/\big(m|\sA_s|\big)$ plays. Since a play of $\pi_m$ consists of $m \le k$ pulls, we have that each $g(m)$ is estimated using $T_s/|\sA_s| \ge T_s/k$ realizations of a sequence of random variables whose expectations have average exactly equal to $g(m)$. 

We now claim that, with probability at least $1-\delta$,
$
    \rstar \in \bigcap_{s=1}^S A_s
$
and
$0 \le  \ghat_s(\mhat_s) - \ghat_s(\rstar) \le 2 C_s$ for all $s=1,\dots,S$.

We prove the claim by induction on $s=1,\dots,S$. We first show that the base case $s=1$ holds with probability at least $1-\delta/S$. Then we show that if the claim holds for $s-1$, then it holds for $s$ with probability at least $1-\delta/S$ over all random events in stage $s$. Therefore, using a union bound over $s=1,\dots,S$ we get that the claim holds simultaneously for all $s$ with probability at least $1-\delta$.

For the base case $s=1$ note that $\rstar \in A_1$ by definition, and thus $0 \le \ghat_1(\mhat_1) - \ghat_1(\rstar)$ holds. Moreover:
$
    \ghat_1(\mhat_1) - g(\mhat_1) \le C_1
$,
$
    g(\rstar) - \ghat_1(\rstar) \le C_1
$, and
$
    g(\mhat_1) - g(\rstar) \le 0
$,
where the two first inequalities hold with probability at least $1-\delta$ because of~(\ref{eq:uniform}). This implies
$
    0
\le
    \ghat_1(\mhat_1) - \ghat_1(\rstar)
\le
    2C_1
$
as required. We now prove the claim for $s > 1$. The inductive assumption \\[1mm]
$
    \rstar \in \sA_{s-1}
$ and
$
    0
\le
    \ghat_{s-1}(\mhat_{s-1}) - \ghat_{s-1}(\rstar)
\le
    2C_{s-1}
$ \\[1mm]
directly implies that $\rstar\in\sA_s$. Thus we have  $0 \le \ghat_s(\mhat_s) - \ghat_s(\rstar)$, because $\mhat_s$ maximizes $\ghat_s$ over a set that contains $\rstar$. The rest of the proof of the claim closely follows that of the base case $s=1$.

We now return to the proof of the theorem. For any $s=1,\dots,S$ and for any $m \in \sA_s$ we have that
\begin{align*}
    g(\rstar) &- g(m)
\le
    g(\rstar) - \ghat_{s-1}(m) + C_{s-1} \quad \text{by~(\ref{eq:uniform})}
\\&\le
    g(\rstar) - \ghat_{s-1}(\mhat_{s-1}) + 3C_{s-1}
\\ &\quad
    \text{by definition of $\sA_{s-1}$, since $m \in \sA_s \subseteq \sA_{s-1}$}
\\&\le
    g(\rstar) - \ghat_{s-1}(\rstar) + 3C_{s-1}
\\ &\quad
    \text{since $\mhat_{s-1}$ maximizes $\ghat_{s-1}$ in $\sA_{s-1}$}
\\&\le
    4C_{s-1} \quad \text{by~(\ref{eq:uniform})}
\end{align*}
holds with probability at least $1-\delta/S$. Hence, recalling that the number of switches between two different policies in $\Pi_{\scK}$ is deterministically bounded by $kS$, the regret of the player can be bounded as follows,
\begin{align*}
   G_T&(\ghost) - G_T(\low)
\\&=
    k^2S + \sum_{s=1}^S \frac{T_s}{|\sA_s|} \sum_{m \in \sA_s} \Big(g(\rstar) - g(m)\Big)
\\&=
    k^2S + T_1
+
    \sum_{s=2}^S \frac{T_s}{|\sA_s|} \sum_{m \in \sA_s} \Big(g(\rstar) - g(m)\Big)
\\&\le
    k^2S + T_1\ + \sum_{i=2}^S 4T_s\sqrt{\frac{k}{2T_{s-1}}\ln\frac{2kS}{\delta}}
\\&=
    k^2S + T_1 + 4\sqrt{k\ln\frac{2kS}{\delta}} \sum_{s=2}^S \frac{T_s}{\sqrt{T_{s-1}}}
\end{align*}
where the $k^2S$ term accounts for the regret suffered in the $kS$ plays where we switched between two policies in $\Pi_{\scK}$ and paid maximum regret due to calibration for at most $k$ steps (as each policy in $\Pi_{\scK}$ is implemented with at most $k$ pulls).
Now, since $T_1 = \sqrt{T}$, $T_s/\sqrt{T_{s-1}} = \sqrt{T}$ and $S = \mathcal{O}\bigl(\ln\ln T\bigr)$, we obtain that with probability at least $1-\delta$ the regret is at most of order
$
k^2\ln\ln T + \sqrt{T} + \sqrt{kT\left(\ln\frac{k}{\delta} + \ln\ln\ln T\right) }
$.
\end{proof}
\section{Learning the ordering of the arms}
\label{Sec:sampling_stage}
%
%
In this section we show how to recover, with high probability, the correct ordering $\mubar_1 > \cdots > \mubar_k$ of the arms. Initially, we ignore the problem of calibration, and focus on the task of learning the arm ordering when each pulls of arm $i$ returns a sample from the true baseline reward distribution with expectation $\mu_i$.


\begin{algorithm}[!t]
\begin{algorithmic}[1]
\Require{Confidence $\delta \in (0,1)$}
\Ensure{A permutation $[1],\ldots,[k]$ of $\scK$.}
\State Let $\sA_1 \equiv \scK$ be the initial set of active arms
\Repeat  \Comment{$r$ indexes the round number}
	\State Sample once all arms in $\sA_r$ \label{line:rr} \Comment{sampling round}
	\State Sort the empirical means $\muhat_{[1],r} \ge\cdots\ge \muhat_{[n],r}$ \label{line:sort}
	\For {$i = 1$ to $|\sA|$}
		\If {${\dt \muhat_{[i],r} + 2\ve_r < \min_{j \in \scK^+_{[i],r}} \muhat_{j,r} }$} \label{line:test1}
			\If {${\dt \muhat_{[i],r} - 2\ve_r > \max_{j \in \scK^-_{[i],r}} \muhat_{[s],r} }$} \label{line:test2}
			\State Remove $[i]$ from $\sA_r$\; \label{line:remove}
			\State Rank before $[i]$ all arms in $\scK^+_{[i],r}$ \label{line:bigger}
			\State Rank after $[i]$ all arms in $\scK^-_{[i],r}$ \label{line:smaller}
		\EndIf \EndIf
	\EndFor
\Until{$|\sA_t| \le 1$} \label{line:stop}
\end{algorithmic}
\caption{(\texttt{BanditRanker})}
\label{Alg:BanditRanker}
\end{algorithm}

\texttt{BanditRanker} (Algorithm~\ref{Alg:BanditRanker}) is an action elimination procedure. The arms in the set $\sA_r$ of active arms are sampled once each (line~\ref{line:rr}), and their average rewards are kept sorted in decreasing order (line~\ref{line:sort}). We use $\muhat_{i,r}$ to denote the sample average of rewards obtained from arm $i$ after $r$ sampling rounds, and define the indexing $[1],\ldots,[k]$ be such that $\muhat_{[1],r} \ge \cdots \ge \muhat_{[k],r}$, where ties are broken according to the original arm indexing.

When the confidence interval around the average reward of an arm $[i]$ is not overlapping anymore with the confidence intervals of the other arms (lines~\ref{line:test1}--\ref{line:test2}), $[i]$ is removed from $\sA_r$ and not sampled anymore (line~\ref{line:remove}). Moreover, the set $\scK^+_{[i],r}$ of all arms $[b] \in \sA_r$ such that $\muhat_{[b],r} \ge \muhat_{[i],r}$ (if any) is ranked before $[i]$ (line~\ref{line:bigger}). Similarly, the set let $\scK^-_{[i],r}$ of all arms $[s] \in \sA_r$ such that $\muhat_{[s],r} \le \muhat_{[i],r}$ (if any) is ranked after $[i]$ (line~\ref{line:smaller}). The algorithm ends when all arms are removed (line~\ref{line:stop}).

The parameter $\ve_r$ determining the confidence interval after $r$ sampling rounds is defined by
\begin{equation}
\label{eq:ve_r}
	\ve_r = \sqrt{\frac{1}{2r}\ln\frac{2kr(r+1)}{\delta}}~.
\end{equation}
The sequence of removed arms can be stored in a binary tree whose root is the first removed arm and whose left (resp., right) leaf contain all arms whose average reward was bigger (resp., smaller) when the first arm was removed. When a new arm is removed, the leaf to which it belongs is split using the same logic that we used for the root. Eventually, all nodes contain a single arm and the in-order traversal of the tree provides the desired ordering.

We introduce the following quantity, measuring the suboptimality gaps between arm that are adjacent in the correct ordering,
\begin{align*}
	\Delta_i = \left\{ \begin{array}{cl}
		\Delta_{1,2} & \text{if $i=1$}
	\\
		\min\big\{\Delta_{i-1,i},\Delta_{i,i+1}\big\} & \text{if $1 < i < k$}
	\\
		\Delta_{k-1,k} & \text{if $i=k$}
	\end{array} \right.
\end{align*}
where $\Delta_{i,j} = \mubar_i - \mubar_j$.

We are now ready to state and prove the main result of this section.
\begin{theorem}
\label{th:ordering}
If Algorithm~\ref{Alg:BanditRanker} is run with parameter $\delta$ on a $k$-armed stochastic bandit problem, the correct ordering $\mu_1 > \cdots > \mu_k$ of the arms is returned with probability at least $1-\delta$ after a number of pulls of order
\begin{equation}
\label{eq:ordering}
	\sum_{i=1}^{k-1} \frac{1}{\Delta_i^2}\ln\frac{1}{\delta\Delta_i}~.
\end{equation}
\end{theorem}
Note that, up to logarithmic factors, the bound stated in Theorem~\ref{th:ordering} is of the same order as the sample used by an ideal procedure that knows $\Delta_1,\ldots,\Delta_k$ and uses the optimal order $1\big/\Delta_i^2$ of samples to determine the position of each arm $i$ in the correct ordering.
\begin{proof}
The proof is an adaptation of \citet[Theorem~8]{ActionElimination}. Using Chernoff-Hoeffding bounds, the choice of $\ve_r$ ensures that
\begin{align}
\nonumber
	\Pr\Big( \exists\,r \ge 1\; \exists i\in\scK \; \big|\muhat_{i,r} - \mu_i\big| > \ve_r \Big)
&\le
	2k \sum_{r\ge 1} e^{-2\ve_t^2 r}
\\ &\le
\label{eq:hoeff}
	\delta~.
\end{align}
If an action $[i]$ is eliminated after $r$ sampling rounds, then it must be that
$
	\muhat_{[b],r} - 2\ve_r > \muhat_{[i],r} > \muhat_{[s],r} + 2\ve_r
$
for all $[b] \in \scK^+_{[i],r}$ and all $[s] \in \scK^-_{[i],r}$. Condition~(\ref{eq:hoeff}) then ensures that, with probability at least $1-\delta$,
$
	\mu_{[b]} > \mu_{[i]} > \mu_{[s]}
$
for all such $b$ and $s$. This implies that the current ordering of $\mu_{[j],r}$ for $j\in\sA_r$ is correct with respect to $[i]$. Since $\ve_r \to 0$, every action is eventually eliminated. Therefore, with probability at least $1-\delta$ the sequence of eliminated arms $i$ and their corresponding sets $\scK^+_{[i],r},\scK^-_{[i],r}$ provide the correct arm ordering. 

We now proceed to bounding the number of samples. Under condition~(\ref{eq:hoeff}), for all $b < i < s$,
\[
	\Delta_{b,i} - 2\ve_r = \big(\mu_b - \ve_r\big) - \big(\mu_i + \ve_r\big) \le \muhat_{b,r} - \muhat_{i,r}~.
\]
Therefore, if $\muhat_{b,r} - \muhat_{i,r} \le 2\ve_r$, then $\Delta_{b,i} \le 4\ve_r$. Recalling the definition~(\ref{eq:ve_r}) of $\ve_r$ and solving by $r = r(b,i)$ we get
\[
	r(b,i) = \scO\left(\frac{1}{\Delta_{b,i}^2}\ln\frac{1}{\delta\Delta_{b,i}}\right)~.
\]
Thus, after $r(b,i)$ sampling rounds, $\muhat_{b,r(b,i)} - \muhat_{i,r(b,i)} > 2\ve_{r(b,i)}$ with probability at least $1-\delta$. Similarly, after $r(i,s)$ sampling rounds, $\muhat_{i,r(i,s)} - \muhat_{s,r(i,s)} > 2\ve_{r(i,s)}$ with probability at least $1-\delta$.

This further implies that after $N_i = \scO\left(\frac{1}{\Delta_i^2}\ln\frac{1}{\delta\Delta_i}\right)$ many sampling rounds, action $i$ is eliminated and not sampled any more.

Re-define the indexing $[1],\ldots,[k]$ so that $\Delta_{[1]} > \cdots > \Delta_{[k]}$. Hence $N_{[1]} < \cdots < N_{[k]}$ by definition. We now compute a bound on the overall number of pulls based on our bound on the number of sampling rounds. With probability at least $1-\delta$, we have that: $k N_{[1]}$ pulls are needed to eliminate arm $[1]$, $(k-1)\big(N_{[2]} - N_{[1]}\big)$ pulls are needed to eliminate arm $[2]$, and so on. Hence, with probability at least $1-\delta$ the total number of pulls needed to eliminate all arms is
\begin{align*}
	\sum_{i=0}^{k-2} & (k-i)\big(N_{[i+1]} - N_{[i]}\big)
\\&=
	k N_{[k-1]} - \sum_{i=0}^{k-2} i \big( N_{[i+1]} - N_{[i]}\big)
\\&=
	k N_{[k-1]} - (k-1)N_{[k-1]} + \sum_{i=0}^{k-2} N_{[i+1]}
\\&=
	N_{[k-1]} + \sum_{i=1}^{k-1} N_{[i]}
\end{align*}
where we set conventionally $N_{[0]}=0$. This concludes the proof of the theorem.
\end{proof}
In order to apply \texttt{BanditRanker} to an instance of B2DEP, we assume that an upper bound $d_0 > \max_i d_i$ be available in advance to the algorithm. This ensures that $\mu_i(d_0) = \mubar_i$ for all $i \in \scK$. In each sampling round $r$, we partition the arms in $\sA_r$ in groups of size $d_0$ and make $2d_0$ pulls for each group by cycling twice over the arms in an arbitrary order. Then, the first $d_0$ pulls in each group are discarded, while the last $d_0$ pulls are used to estimate the expectations $\mu_i$ (when $d_0$ does not divide $|\sA_r|$ we can add to $\sA_r$ arms that were already removed, or arms from previous groups, just for the purpose of calibrating). The sample size bound~(\ref{eq:ordering}) remains of the same order (because the extra pulls only add a factor of two).


\section{Experiments}
\label{Sec:Experiments}
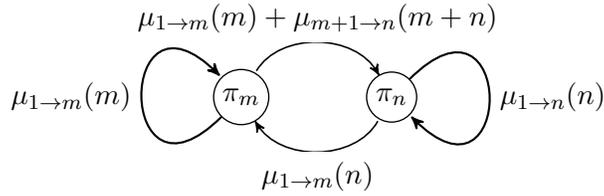
\begin{figure}[ht]
    \centering
    \begin{tikzpicture}[scale=0.5]
\grProb{$\pi_m$}{$\pi_n$}{\mu_{1 \to m}(m) + \mu_{m+1 \to n}(m+n)}{\mu_{1\to m}(n)}{\mu_{1\to m}(m)}{\mu_{1 \to n}(n)}
\end{tikzpicture}
    \caption{Transitions between policies $\pi_m$ and $\pi_n$ assuming $n > m$, where the notation $\mu_{m \to n}(d)$ stands for $\mu_m(d)+\cdots+\mu_n(d)$. The expected reward obtaining by switching between policies is different from the expected reward obtaining by cycling over the same policy.}
    \label{fig:switchGraph}
\end{figure}

In this section we present an empirical evaluation of our policy \low\ in a synthetic environment with Bernoulli rewards. In order to study the impact the switching cost on ranking policies when the suboptimality gap is small, we also define a setting in which there are two distinct ranking policies that are both optimal ---see Figure~\ref{fig:switchGraph}.

We plot regrets against the policy \ghost. Our policy \low\ is run without any specific tuning (other than the knowledge of the horizon $T$) and with $\delta$ set to $0.1$ in all experiments. The benchmark \piucb\ consists of running UCB1 ---with the same scaling factor as in the original article by \citet{UCB1}--- over the class $\Pi_{\scK}$ of ranking policies, where calibration is addressed by rolling out twice each ranking policy selected by UCB1 and using only the second roll-out to compute reward estimates. Since both \low\ and \ghost\ are run over $\Pi_{\scK}$, we implicitly assume that \texttt{BanditRanker} successfully ranked the arms in a preliminary stage.

\begin{figure}[ht]
    \centering
    \includegraphics[height=4cm]{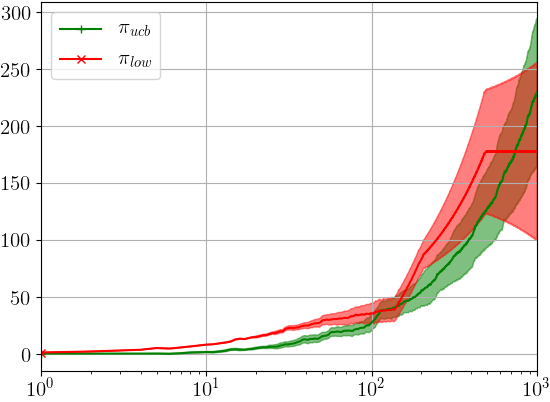}
    \caption{Comparing regrets of \low\ and \piucb\ against \ghost\ with $7$ arms and baseline expectations $0, 1/3, 2/3, 4/5, 13/15, 14/15, 1$ and $f(\tau) = (0.999)^{\tau}$. A unit cost is charged for switching between ranking policies. Curves are averages of $5$ runs each using a different sample of delays $d_1,\ldots,d_7$ uniformly drawn from $\{1,\ldots,6\}$. We plot expectations of sampled arms rather than realized rewards.}
    \label{fig:std_plot}
\end{figure}
Figure~\ref{fig:std_plot} shows that when the gap between the best and the second best ranking policy is not too small ($0.1$ on average in these experiments), then \piucb\ is competitive against \low\ even in the presence of unit switching costs. This happens because, in order to minimize the number of switches, \low\ samples a suboptimal policy more frequently than $\piucb$. Although this oversampling does not affect the distribution-free regret bound of \low, it hurts performance unless the suboptimality gap is small enough to cause the switching costs to prevail, a case which is addressed next. Note also that \low\ eventually stops exploration because all policies but one have been eliminated, while \piucb\ keeps on exploring, albeit at a logarithmic rate.

\begin{figure}[ht]
    \centering
    \includegraphics[height=4cm]{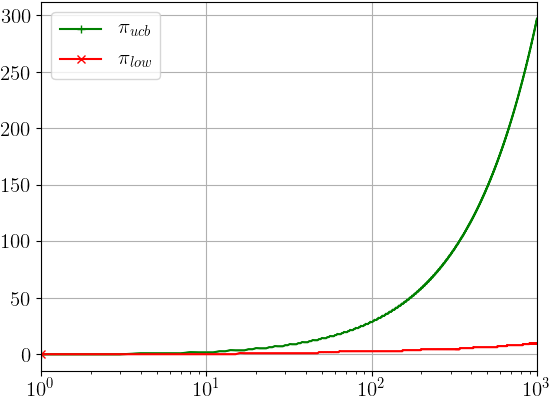}
    \includegraphics[height=4cm]{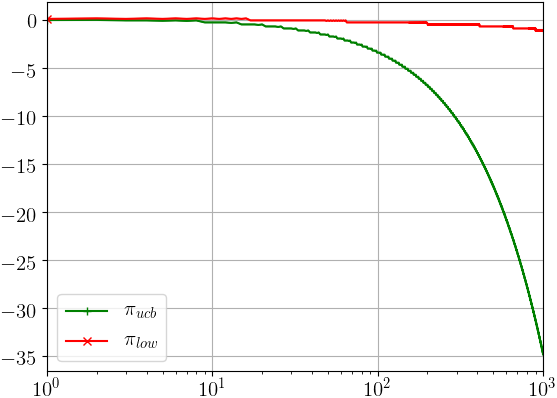}
    \caption{Comparing regrets of \low\ and \piucb\ against \ghost\ with $2$ arms such that $g(1) = g(2)$ with unit cost charged for switching between the two policies (upper part) and without any cost for switching (lower part).}
    \label{fig:SWITCH_COST}    
\end{figure}

In the second experiment we consider two arms with $\mu_1 = 1$, $f(1) = 0.3$, $f(2) = 0.25$, $d_1=d_2=2$, and $\mu_2$ chosen so that $g(1) = g(2)$ to simulate a vanishing suboptimality gap between $\pi_1$ and $\pi_2$. Figure~\ref{fig:SWITCH_COST} (upper part) shows that \low\ performs better than \piucb\ due to its low switch regime. On the other hand, Figure~\ref{fig:SWITCH_COST} (lower part) shows that when the switching cost is zero, switching between two good policies becomes more advantageous than using a single good policy, and the regret of both \piucb\ and \low\ becomes negative (in this case \piucb, which has no control over the number of switches, outperforms \low). The reason for this advantage is explained by Fact~\ref{Th:Switching} below (proof in the supplementary material), see also Figure~\ref{fig:switchGraph}.
\begin{fact}\label{Th:Switching}
    If an instance of B2DEP admits two optimal ranking policies, then consistently switching between these two policies achieves an average expected reward higher than sticking to either one.
\end{fact}
To summarize, the experiments confirm that, in the presence of switching costs, \low\ works better than \piucb\ only when the suboptimalty gap is very small. The advantage of \low\ over \piucb\ is however reduced by the fact that switching between two good policies is better than consistently playing either one of the two (Fact~\ref{Th:Switching}). Note also that \low\ stops exploring because $T$ is known. This preliminary knowledge can be dispensed with using a doubling trick, or some more sophisticated method. Also, it would be interesting to design a method that achieves the best between the performance of \piucb\ and \low, according to the size of the suboptimality gap. 

\section{Conclusions}
Motivated by music recommendation in streaming platforms, we introduced a new stochastic bandit model with nonstationary reward distributions. To cope with the NP-hardness of learning the optimal policy caused by nonstationarity, we introduced a restricted class of ranking policies approximating the optimal performance. We then proved sample and regret bounds on the problem of learning the best ranking policy in this class. One of the main problems left open by our work is that of deriving more practical learning algorithms, able to simultaneously learn the ranking of the arms and the best cutoff value $\rstar$, while minimizing their regret with respect to the best ranking policy.

\subsection*{Acknowledgments}
Nicol\`{o} Cesa-Bianchi acknowledges partial support by the Google Focused Award \textsl{Algorithms and Learning for AI} (ALL4AI) and by the MIUR PRIN grant \textsl{Algorithms, Games, and Digital Markets} (ALGADIMAR).
\bibliographystyle{plainnat}
\bibliography{main}

\clearpage
\onecolumn
\begin{center}
\setcounter{section}{0}
\textbf{\large Supplementary Material for Bandits with Delay-Dependent Payoffs}
\end{center}

\section{Proof of Theorem \ref{Th:Hardness}}
\begin{proof}
Given an instance $\ell_1,\dots,\ell_n$ of PMSP, we construct a B2DEP instance with $|\mathcal{K}|=n+1$ arms such that $d_i = \ell_i-1$ and $\mu_i = 1$ for all $i=1,\dots,n$, $\mu_{n+1}=0$, and $f \equiv 1$. The long-term average reward for a periodic policy in this setting is
\[
	\sum_{i=1}^n \frac{1}{N_i} \sum_{j=1}^{N_i} \frac{\Ind{t_{i,j} > d_i}}{t_{i,j}}
\]
where $N_i$ is the number of times the policy plays arm $i$ in a period and $t_{i,j}$ is the number of time steps between when arm $i$ was played for the $j$-th time in the cycle and the last time it was played (in the same cycle or in the previous cycle, excluding the transient). Clearly, if the PMSP instance has a feasible schedule, then we can design a bandit policy that replicates that schedule (playing arm $n+1$ at all time steps where no machines are scheduled for maintenance). The long-term average reward of this policy is at most $\sum_{i=1}^n\frac{1}{d_i+1}$. Moreover, if we have a periodic bandit policy with long-term average reward exactly equal to $\sum_{i=1}^n\frac{1}{d_i+1}$, this means that each arm $i=1,\dots,n$ is eventually played after exactly $d_i+1 = \ell_i$ rounds.Indeed, the only way to have
\[
	\frac{1}{N_i} \sum_{n=1}^{N_i} \frac{\Ind{t_{i,j} > d_i}}{t_{i,j}} \ge \frac{1}{d_i+1}
\]
is by setting $t_{i,j} = d_i+1$ for all $j=1,\dots,N_i$. 
\end{proof}

\section{Proof of Fact \ref{Th:Switching}}
\begin{proof}
We use the following notation: $\mu_{m \to n}(d)$, where $n > m$, stands for $\mu_m(d)+\cdots+\mu_n(d)$. Consider two optimal ranking policies $\pi_m$ and $\pi_n$ with $n > m$. Then $g(m) = g(n)$, where $g(n) = \frac{1}{n}\mu_{1\to n}(n)$ and similarly for $g(m)$. The expected total reward of playing $\pi_m$ after $\pi_n$ is $\mu_{1\to m}(n)$, and the expected total reward of playing $\pi_n$ after $\pi_m$ is $\mu_{1 \to m}(m) + \mu_{m+1 \to n}(m+n)$. We want to prove
\begin{align*}
    \frac{\mu_{1\to m}(n) + \mu_{1 \to m}(m) + \mu_{m+1 \to n}(m+n)}{m+n}
\ge
    \frac{\mu_{1 \to m}(m)}{m}~.
\end{align*}
Rearranging gives
$
    \mu_{1\to m}(n) + \mu_{m+1 \to n}(m+n)
\ge
    \frac{n}{m} \mu_{1 \to m}(m)
$.
Since $\frac{1}{n}\mu_{1\to n}(n) = \frac{1}{m}\mu_{1\to m}(m)$, we have
\[
    \mu_{1\to m}(n) + \mu_{m+1 \to n}(m+n)
\ge
    \mu_{1 \to n}(n)~.
\]
Observing that $\mu_{1 \to n}(n) = \mu_{1\to m}(n) + \mu_{m+1\to n}(n)$, the above is equivalent to
\begin{align*}
    \mu_{m+1 \to n}(m+n)
\ge
    \mu_{m+1\to n}(n)
\end{align*}
which is always true since in our model expected rewards are non-decreasing with delays.
\end{proof}

\end{document}